# LCV2: An Efficient Pretraining-Free Framework for Grounded Visual Question Answering

Yuhan Chen[1], Lumei Su[1,2,*], Lihua Chen[1], Zhiwei Lin[1]

[1] School of Electrical Engineering and Automation, Xiamen University of Technology, Xiamen, China;

[2] Xiamen Key Laboratory of Frontier Electric Power Equipment and Intelligent Control, Xiamen, China;

* Correspondence: sulumei@163.com

**Abstract**. In this paper, the LCV2 modular method is proposed for the Grounded Visual Question Answering task in the vision-language multimodal domain. This approach relies on a frozen large language model (LLM) as intermediate mediator between the off-the-shelf VQA model and the off-the-shelf visual grounding (VG) model, where the LLM transforms and conveys textual information between the two modules based on a designed prompt. LCV2 establish an integrated plug-and-play framework without the need for any pre-training process. This framework can be deployed for VQA Grounding tasks under low computational resources. The modularized model within the framework allows application with various state-of-the-art pre-trained models, exhibiting significant potential to be advance with the times. Experimental implementations were conducted under constrained computational and memory resources, evaluating the proposed method's performance on benchmark datasets including GQA, CLEVR, and VizWiz-VQA-Grounding. Comparative analyses with baseline methods demonstrate the robust competitiveness of LCV2.

## 1. Introduction

With the refinement and maturation of Computer Vision (CV) and Natural Language Processing (NLP) technologies in the field of deep learning, research and applications focusing on a single modality can no longer satisfy the growing demands for further advancements in artificial intelligence. The fusion of multimodal visual and language information has become an inevitable development in the field of deep learning. The emergence of transformer-related work, denoted as Transformer [1], BERT [2], ViT [3], et al. has notably propelled the progress of multimodal technologies in visual-language integration. In recent years, multimodal visual-language research has given rise to numerous downstream tasks, including Visual Question Answering (VQA) [4], Visual Entailment (VE) [5], Visual Commonsense Reasoning (VCR) [6], and others.

This work focuses on the task of multimodal Visual Question Answering (VQA) and the grounding of visual cues related to questions and answers. Visual Question Answering, as a cross-modal downstream task, enables systems to comprehend information from both images and text, generating natural language answer about images. It holds broad potential applications, including intelligent assistants, image searches, and more extensive human-computer interaction systems. Early approaches to VQA primarily involved the joint embedding of visual and textual features, as seen in the work of [16-19]. Subsequently, with the introduction of the transformer [1] architecture, the extraction of visual and textual features, as well as the fusion alignment process, has been concentrated on employing the transformer architecture. Examples of such advancements in this stage include the works of ViLBERT [7], Visual-BERT [8], and Oscar [9]. These multimodal models have demonstrated significant improvements in performance.

As VQA systems mature, there is a shift away from merely eliciting textual responses. Some works, such as [10-13], enable models not only to generate textual responses based on visual information but to explicitly provide corresponding visual cues, such as bounding boxes (bbox), image segmentation, or heatmap representations. This has given rise to the VQA grounding task, as illustrated in Figure 1. This task finds applications in visual navigation and precise localization for individuals with visual impairments. Additionally, some works leverage models to provide visual evidence, ensuring the correctness and interpretability of the responses. Early contributions to VQA grounding tasks include the representative work MAC-Caps [10], which utilizes traditional Convolutional Neural Networks (CNN) and Recurrent Neural Networks (RNN) to extract visual and textual features separately. The fusion of these features occurs in the inference module, producing the answer text and an attention map. The attention map represents the answer grounding. Subsequently, the advantages of the Transformer framework facilitated the development of DaVI [11]. DaVI employs the Transformer architecture to construct visual and language encoders. It generates an answer text and a grounding mask through language decoders and language-based visual decoders, respectively.

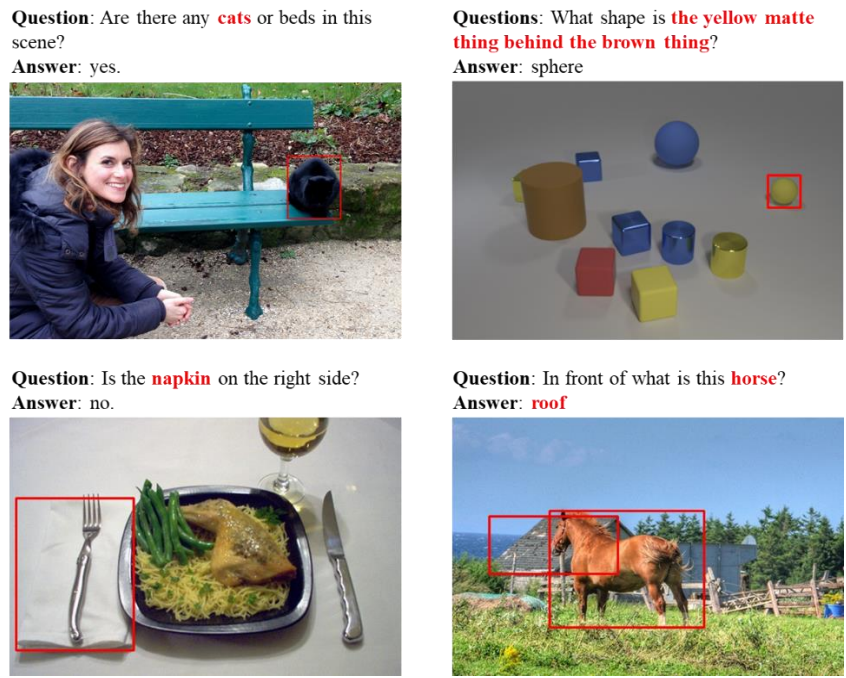

**Figure** 1. The figure illustrates an intuitive visualization of the VQA Grounding task. In this VQA grounding multimodal system, besides providing textual answers to questions about visual content, it is essential to distinctly ground the positions of objects mentioned in the text within the corresponding visual context.

While extensive progress has been made in research on VQA grounding tasks, addressing the challenges associated with open-world visual-language multimodal models remains a significant obstacle, since the substantial computational power, storage resources, and extensive graph-text training data are required during additional pre-training stages. This situation severely hinders institutions and enterprises with limited computational and data resources from effectively engaging in VQA grounding task applications. Instead, a powerful strategy in scenarios with constraints on computational power, storage, and training data resources is the adoption of pretraining-free VQA grounding systems.

Particularly noteworthy is the emergence of universal pre-trained models in recent years and the revolutionary developments in Large Language Models (LLMs). These advancements have made it feasible to construct frameworks for pretraining-free VQA grounding systems.

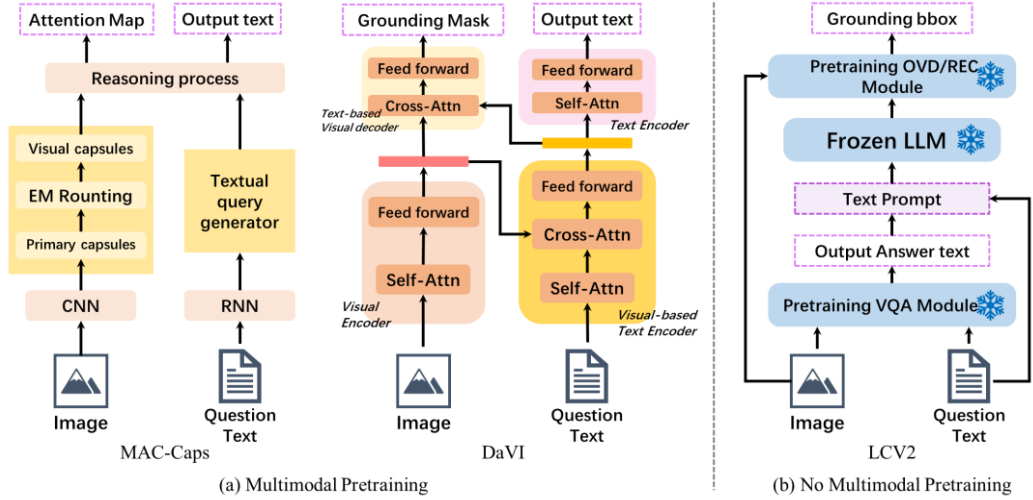

Figure 2. Demonstrating the disparities between our approach and other representative baseline methods. (a) Represents methods requiring multimodal pretraining, where MAC-Caps relies on conventional CNN and RNN to extract visual and textual features separately. The DaVI method adopts a Transformer-based architecture for encoding and decoding multimodal features of images and text. (b) Represents our proposed approach that is free from multimodal pretraining. It is based on the LLM model to connect the VQA module and OVD/REC module, thereby constructing an integrated system for the VQA Grounding task.

Therefore, addressing the aforementioned challenges in the VQA grounding task, we propose a pretraining-free modular approach. Leveraging the powerful text understanding, reasoning, and generative capabilities of Large Language Models (LLM), we employ it as an intermediate mediator between a pre-trained VQA downstream task model and a pre-trained visual grounding (VG) model (including Open-Vocabulary Object Detection (OVD) and Referring Expression Comprehension (REC) in this paper) for transferring and transforming relevant textual information. We term our approach LCV2 (The **L**LM **C**onnects the **V**QA module and **V**G module). Figure 2 illustrates the differences between our modular approach and the baseline methods MAC-Caps [10] and DaVI [11], which require pre-training processes. Specifically, in our approach, the textual question and corresponding image information are initially fed into a visual-text multimodal VQA pre-trained model. The model generates textual responses based on visual information. Subsequently, this textual response and the question text are further fed into the LLM, where a designed prompt prompts the LLM to transform the Q&A text into descriptive caption text. Following this transformation, the caption text and visual content are further fed into the pre-trained OVD/REC model, which provides explicit bounding box (bbox) annotations for the described objects. This framework achieves VQA grounding by providing both textual answers to questions and visually explicit bounding box indications.

Our modular approach avoids the substantial computational and data costs associated with the pretraining stage. By integrating state-of-the-art pre-trained models, we achieve a plug-and-play multimodal VQA grounding system. Moreover, our design of modular integration enables the utilization of most advanced pre-trained models and methods from the computer vision and natural language processing domains, maximizing model performance and harnessing the strengths of the framework.

This paper summarizes the contributions of our work as follows:

I) We propose LCV2, a modular integrated approach designed for the Visual Question Answering

(VQA) grounding task. LCV2 connects an off-the-shelf visual question answering module and an off-the-shelf visual grounding (VG) module through a frozen Large Language Model with powerful text semantic transformation capabilities, establishing a modular framework.

II) LCV2 circumvents the pretraining stage, eliminating the need for extensive computational and training data resources. This enables the implementation of VQA grounding under conditions of low computational power and low cost. Moreover, the model is capable of integrating state-of-the-art pre-trained models and methods, allowing for performance improvement in tandem with advancements in the community.

III) Experimental results on benchmark datasets, including GQA, CLEVR, and VizWiz-VQA-Grounding, demonstrate the competitiveness of our approach compared to other baseline methods.

## 2. Related Work

### 2.1 Visual Question Answering and VQA Grounding

Early Visual Question Answering (VQA) efforts focused on implementing joint embedding methods, where the extraction of visual features and textual features relied on Convolutional Neural Networks (CNNs) [14] and Recurrent Neural Networks (RNNs) [15], respectively. The fusion of these cross-modal features was achieved through simple mechanism combinations. Malinowski et al. [16] exemplified this approach, employing semantic segmentation for image processing and utilizing Bayesian algorithms to handle both images and questions. Building upon Malinowski et al.'s work, Ren et al. [17] introduced improvements by sending features generated by the encoder to a classifier to obtain answers. Yu et al. [18] and Ben et al. [19] applied various multimodal bilinear pooling methods to combine image features from spatial grids with text features from questions.

The introduction of the Transformer [1] architecture ushered in a new era for the development of cross-modal approaches. Influenced and inspired by the work of BERT [2] in the field of Natural Language Processing, the visual-text multimodal domain also began to explore the "pretraining-finetuning" paradigm for large models. Numerous pretraining models, based on the Transformer architecture, emerged in the visual-text multimodal domain. Examples include ViLBERT [7], Visual-BERT [8], Oscar [9], BLIP [20], BLIP-2 [21], Flamingo [22], and others. The VQA task is regarded as a downstream task in the fine-tuning training of cross-modal large models, and has gained important development and progress in this stage.

With the extensive research on Visual Question Answering (VQA) tasks and the continued maturation of cross-modal models, there was no longer restricted VQA systems to answering text-based questions based on visual content. Recent efforts have focused on exploring the capability of models to provide question responses while simultaneously grounding visual cues. The MAC network [12], for instance, addresses this task by decomposing questions into a series of attention-based reasoning steps, using attention heatmaps to indicate visual cues. MAC-Caps [10] introduces a query-based selection mechanism of capsule features in the visual capsule module, further enhancing performance. Aurora [11] proposes Dual Visual-Linguistic Interaction (DAVI) and achieved first place in the answer grounding track of the 2022 VizWiz Grand Challenge. To further facilitate research on VQA grounding tasks, some VQA benchmark datasets have provided available grounding labels. Examples include GQA [23], CLEVR [24], VCR [25], VizWiz-VQA-Grounding [26], among others.

### 2.2 Open-Vocabulary Object Detection and Referring Expression Comprehension

The task of Open-Vocabulary Object Detection (OVD) involves language-based generalization for detecting objects of any category within visual content. It entails understanding captions information and accurately indicating the bounding box positions of objects involved in the visual content. In earlier years, OV-DETR [27] utilized a diverse set of image-caption pairs to enhance the model's detection capabilities for unknown classes. In recent years, with the widespread exploration of the Transformer [1] architecture and inspiration from contrastive learning methods applied in CLIP [28], significant developments have occurred in open-vocabulary object detection methods. The work of GLIP [29] unifies object detection and phrase grounding tasks, enabling models to simultaneously learn from both types of data and leverage a large number of network image-text pairs for self-training. DetCLIP [30] introduces a novel parallel training framework to better utilize heterogeneous datasets for training and simultaneously constructs an additional knowledge base to provide implicit relationships between classes. Grounding DINO [31] argues that previous works have overlooked the crucial scenario of open-set object detection—evaluation in the task of Referring Expression Comprehension (REC). It proposes a closely integrated solution, prompting models to detect any target mentioned in the input text, including category names and referring expressions. Xie et al. [32] distinguish Open-Vocabulary Object Detection and Referring Expression Comprehension (REC), introducing a novel concept called Described Object Detection (DOD) as a superset encompassing OVD and REC. This elevation positions them in a more practical context.

## 2.3 Large Language Model

The emergence of the Transformer architecture provided the architectural foundation and feasibility for large language models, fostering revolutionary progress in Natural Language Processing. The introductions of GPT-1 [33] and BERT [2] marked the onset of the era of pretrained language models, with "pretraining-finetuning" becoming a new paradigm for training large language models. Subsequent developments witnessed a continuous increase in model parameters, giving rise to models such as GPT-2 [34], T5 [35], and OPT [36]. Flan-T5 [37] demonstrated strong generalization capabilities by implementing instruction fine-tuning on a massively scaled task. ERNIE 3.0 [38] introduced an extensive knowledge graph, proposing strategies for parallel pretraining of large-scale unlabeled text and knowledge graph. Llama 2 [39], leveraging an optimized autoregressive Transformer, completed pretraining on a dataset of 20 trillion tokens, demonstrating advanced model performance. The releases of GPT-3 [40] and GPT-3.5 [41] generated widespread acclaim worldwide, showcasing outstanding performance in various language tasks, including text translation, content creation, logical inference, and code generation. GPT4.0 [42], surpassing GPT3.5, exhibits significant advancements with cross-modal understanding and generation capabilities, standing as one of the most advanced large models to date. Large language models such as Qwen [43], PaLM [44], and ChatGLM [45] have also exhibited powerful competitive capabilities in the model landscape.

## 3. Method

### 3.1 Background

**Transformer**. The Transformer architecture, a deep learning model based entirely on the self-attention mechanism, was originally proposed in the field of natural language processing. Subsequent endeavors gradually extended its application to various sequence-to-sequence tasks and even visual tasks. Presently, the prevailing paradigm in visual-image multimodal research widely adopts the Transformer framework. This choice is motivated by its superior performance in handling long-range dependencies and parallel computations, as well as its ease of adaptation to diverse types of input data. The pre-trained models for various multimodal tasks involved in our proposed modular approach are all implemented based on the Transformer architecture. A key feature and advantage of the Transformer lie in its use of the self-attention mechanism to establish relationships between elements within a sequence. The mathematical process of the self-attention mechanism can be described by Equation (1).

$$\text{Att}(X_q W_Q, X_k W_K, X_v W_V) = \text{softmax}\left(\frac{X_q W_Q \cdot (X_k W_K)^T}{\sqrt{d}}\right) X_v W_V \tag{1}$$

Where $\{W_Q, W_K, W_V\}$ denote trainable linear transformation matrices, and $\{X_q, X_k, X_v\}$ represent input parameters. The self-attention mechanism calculates attention coefficients for each element in the sequence with respect to other elements, capturing long-range dependencies within the sequence. This addresses limitations in handling long-distance dependencies that are encountered by traditional CNNs and RNNs.

**LoRA**. The paradigm of "pretraining-finetuning" has become prevalent for training large models; however, full-parameter fine-tuning incurs significant computational and storage costs. Parameter-efficient fine-tuning (PEFT) [46] achieves approximate or even equivalent results to full fine-tuning by updating only a portion subset of or an additional part of the parameters. It has gained favor in the context of fine-tuning large-scale models across modalities and large language models. The LoRA [47] of PEFT approach employs additional updatable low-rank matrices as bypass matrices to emulate full-parameter fine-tuning. This is achieved based on the inherent low-rank characteristics. The fine-tuning method is utilized in the experimental sections denoted as Section 4.5 and Section 4.6, fine-tuning a pre-trained Visual Question Answering (VQA) model for applicability across different dataset ranges.

The forward propagation process after LoRA fine-tuning is expressed by Equation (2). We denote $W_{tr}$ as an additional low-rank matrix, and $W_{pre}$ represents the frozen pre-trained matrix, i.e., $W_{pre} = \{W_Q, W_K, W_V\}$. In our implemented application of fine-tuning, the additionally trainable bypass matrix $W_{tr}$ is residual-connected to $W_Q$, $W_K$, or $W_V$ of the self-attention mechanism, as depicted in Equation (1).

$$\begin{cases} Y = W_{pre}X + W_{tr}X = W_{pre}X + W_B W_A X \\ W_B \in \mathbb{R}^{\alpha \times \beta}, W_A \in \mathbb{R}^{\beta \times \gamma}, and\ \beta \ll \min(\alpha, \gamma) \end{cases} \quad (2)$$

Where $W_A$ and $W_B$ represent the dimension reduction matrix and dimension expansion matrix, respectively, while $X$ denotes the input parameters for the forward channel of the pre-trained matrix.

### 3.2 Modular Framework

The pre-training-free modular approach we propose is named LCV2. It leverages the Large Language Model (LLM) as an intermediate mediator to connect the VQA model and OVD/REC model. Built upon the language text representations between frozen LLM transformation models, this unified framework system achieves VQA grounding tasks. The pipeline of our modular approach is illustrated in Figure 3. Visual content and associated question text are initially processed and inferred by the frozen pre-trained VQA model, generating predicted answer text. The frozen pre-trained VQA model is interchangeable, offering options such as BLIP [20], Lens [48], GIT [49], etc., due to the integrated modular design of our framework. The question text and predicted answer text are collected, and based on this text, prompts for the LLM are designed and fed to the frozen pre-trained large language model. Guided by the prompts, the LLM transforms the Q&A text into descriptive captioning. The pre-trained Flan-T5 open-source model is employed as the foundational model for our LLM in this paper. Subsequently, the descriptive captions and visual content are further fed into the frozen pre-trained OVD/REC model. The OVD/REC module in this paper is based on the pre-trained Grounding DINO, known for achieving state-of-the-art performance in phrase grounding and referring expression comprehension tasks. The model is capable of performing phrase grounding or referring expression grounding tasks based on the relevant caption text, generating bounding box annotations indicating the location of the referred object. Above-mention, our modular approach achieves question-answering on visual content and locates object regions within the visual content.

Our approach employs a design strategy of integrating off-the-shelf models tailored for different downstream tasks, thereby eliminating cross-modal pre-training and achieving a plug-and-play VQA grounding system. Each frozen pre-trained model can be freely replaced and recombined, with the potential and capability to leverage state-of-the-art advancements in the fields of multimodal and natural language processing. In the subsequent sections of the paper, key points of the modules will be introduced, providing an overview of the main modular components within the framework.

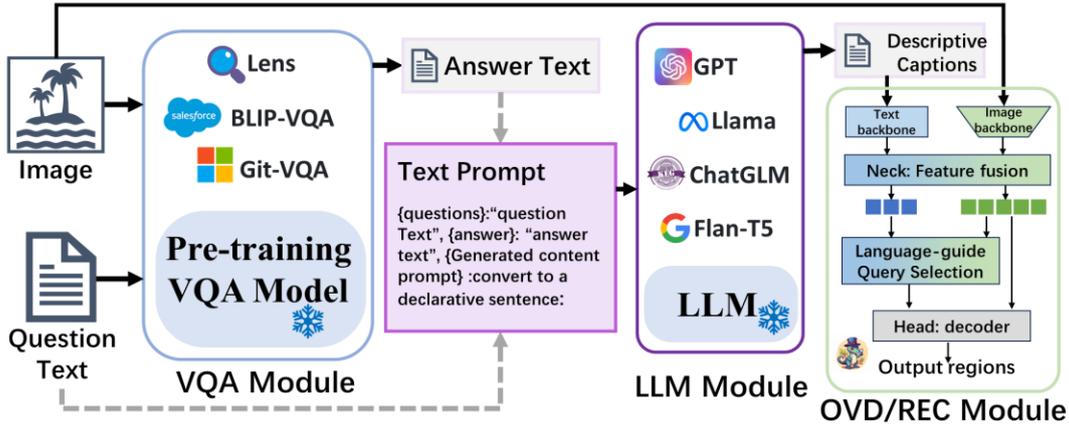

**Figure** 3. Pipeline diagram of LCV2. Visual content and textual questions are first fed into the VQA module to generate predicted answers. The predicted answer text, along with the original question text, is further collected and designed as a prompt to guide the large language model module in generating descriptive caption text. Finally, the descriptive captions and visual content are fed into the OVD/REC module, producing visual bounding boxes that indicate the visual clues for VQA.

### 3.3 LCV2 Components

LCV2 comprises three primary modular models: the VQA module, the LLM module, and the OVD/REC module. Each module is designed according to its respective pre-training task, accomplishing distinct objectives, and integrated into a unified VQA grounding system.

### 3.3.1 VQA Module

In this paper, we primarily replaced and examined models within the VQA module, including BLIP-VQA, Lens, and GIT-VQA. In the selection of models, we primarily considered lightweight model parameters to facilitate rapid inference in scenarios with limited computational and storage resources. In the following sections, we provide an overview of the adopted models:

**BLIP-VQA**. BLIP-VQA [20] is a downstream task model for visual question answering derived from BLIP. It comprises an image encoder based on Visual Transformer (ViT), a text encoder grounded in image content for feature fusion and alignment, and an answering decoder. The pre-training phase involves tasks such as Image-Text Contrast (ITC), Image-Text Matching (ITM), and Language Modeling (LM). This state-of-the-art pre-trained model allows us to input both image content and question text to generate corresponding responses. In numerous subsequent experimental validations in this paper, the BLIP-VQA-Large version is employed for visual question answering tasks due to its demonstrated state-of-the-art performance in the VQA domain.

**Lens**. The Lens [48] framework is a pre-training-free modular approach to visual question answering proposed by the Stanford University team. It utilizes a visual description model, primarily based on CLIP and BLIP, to generate detailed labels, attributes, and captions related to visual content. Ultimately, it employs a large language model as the inference module, guiding the LLM to generate the final responsive text through prompt design. However, Lens heavily relies on the visual description model, with the granularity of visual descriptions determining the model's grasp and understanding of visual information.

**GIT-VQA**. The GIT-VQA [49] model not only provides answers to general visual questions but also exhibits robust performance in recognizing text and symbols within images. It can be applied in scenarios involving questions about images containing simple text and symbols. GIT-VQA is implemented based on a straightforward architecture, where the image undergoes processing by a visual encoder. Subsequently, it is tokenized and embedded along with text and fed into a text decoder composed of multi-head attention layers and feedforward layers. The model's simplified architecture achieves high-performance rapid inference.

In LCV2, visual content and question text are fed into the pre-trained VQA module to generate corresponding predicted answers. The predicted answer text, along with the collected, unaltered question text, is then passed to the Large Language Model (LLM) for descriptive transformation. Considering that the accuracy of the VQA module directly impacts the method's performance in handling VQA grounding tasks, some pre-trained VQA models exhibit a significant decline in performance when applied to entirely new styles of datasets. Retraining these VQA models becomes a reliable measure to ensure model performance. In subsequent experimental validations, we additionally conducted LoRA parameter-efficient fine-tuning training on the VQA module to investigate the impact of introducing new knowledge on overall performance.

### 3.3.2 LLM Module

Due to the modularity of our approach, the Large Language Model (LLM) module is also implemented as plug-and-play, allowing it to be freely selected or replaced with other SOTA LLMs such as the ChatGPT series, LLaMa series, ChatGLM, etc. In this paper, we employ the frozen Flan-T5-large to constitute the foundational LLM module in our framework. This choice is based on considerations of lightweight model parameters, and its excellent performance across various Natural Language Processing (NLP) tasks, suitability for non-complex text tasks in practical applications, and the lightweight nature of model parameters, ensuring real-time inference implementation.

We design a complete prompt to guide the Large Language Model (LLM) in text content generation. The predicted answer text obtained from the VQA module is combined with the original question text to construct the prompt. The complete prompt is formatted as follows: {questions:} "question text" {answer:} "reply text" {Generated content prompt} "convert to a declarative sentence:".

The LLM is prompted to generate descriptive captions regarding the question text and predicted answer text. These descriptive captions are further utilized for the text prompt in open-world object detection tasks. The LLM module serves as an intermediate mediator, extracting textual information, transforming text descriptions, and connecting the VQA module and OVD/REC module.

### 3.3.3 OVD/REC Module

In LCV2, the OVD/REC module is employed to interpret the descriptive captions generated by the LLM, annotating bounding boxes for objects referred to within these descriptions. In our approach, the OVD/REC module is realized based on the Grounding DINO pre-trained model, which possesses dual functionalities of open-vocabulary object detection and referring expression comprehension.

As illustrated in Figure 4, the Grounding DINO model pipeline involves three stages of fusion for visual and text information. Initially, features are separately extracted from the visual backbone and text backbone, and then fused in the Neck section. The query in the Head section is initialized in the Language-guide Query Selection part and further integrated with text features. Finally, the fused cross-modal information in the Head section is employed as a cross-modal decoder to predict bounding box information.

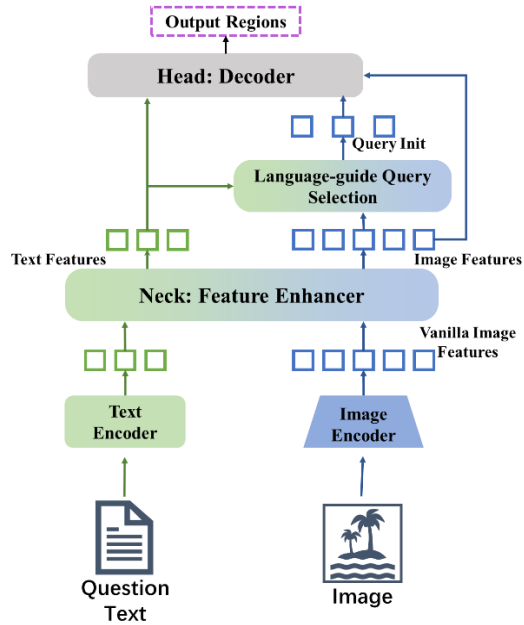

**Figure** 4. Pipeline diagram of the Grounding DINO model: Visual and textual content features are extracted separately by the image encoder and text encoder, with feature fusion performed in the neck section. The Language-guide query selection is employed for initializing the Transformer's query, and the predicted bounding box is finally output in the Head section.

In LCV2, the OVD/REC module utilizes and evaluates both the Grounding DINO-T version model and the Grounding DINO-B version model. The former is pre-trained on the O365, GoldG, and Cap4M datasets, while the latter undergoes pre-training on more extensive datasets such as COCO, O365, GoldG, Cap4M, OpenImage, and others. The models possess extensive general knowledge in the open world, demonstrating competence in visual grounding-related tasks.

## 4. Experiment and Discussions

### 4.1 Datasets

We conducted experiments to validate the performance of LCV2 on the validation sets of publicly available datasets GQA and CLEVR, comparing its performance with common baseline methods. Additionally, to assess and verify the performance of LCV2 in answer grounding applications for visually impaired individuals, we conducted experiments on the test set of the publicly available VizWiz-Answering-Grounding dataset, comparing it with some SOTA methods. In this section, the paper will provide an overview of the publicly available dataset information involved in the experiments.

**GQA**. The GQA dataset [23] is a collection developed by the Stanford University Manning group, focusing on real-world visual reasoning and question answering. Its inception aims to address the evident language bias issues present in previous VQA datasets by employing a more intricate question generation approach. In contrast to the VQA 2.0 dataset, it places a greater emphasis on training and validating the reasoning capabilities and compositional behavior of VQA models. The GQA dataset introduces scene graph information, associating each image with a scene graph describing objects, object bounding box information, properties, and more related to the visual content. This design imparts realism to the questions and facilitates the training and validation of VQA grounding tasks using the dataset. The GQA dataset encompasses 22.66 million questions and over 110,000 images, demanding models to possess extensive inference steps and reasoning skills to answer questions related to visual content.

**CLEVR**. The CLEVR dataset, introduced by Johnson et al. [24], is a synthetic diagnostic dataset widely employed for training and testing various visual-textual reasoning capabilities in VQA or VQA grounding task models. The visual image content in the dataset does not depict real-world scenes; instead, it consists of abstracted stereometric geometry and objects. This design requires models to rely on visual reasoning to perform various VQA tasks, minimizing potential biases and implicit clues while aiming for minimal dataset bias. The dataset's question texts are generated programmatically, covering functional categories such as counting, quantity comparison, existence, color inquiry, material inquiry, size, etc. These categories depict a wide range of textual-visual reasoning types involved in model training and testing. The dataset provides a total of 100,000 rendered images and over 86,000 question-answer text pairs across the training, validation, and test sets. Additionally, detailed scene graphs are provided, containing information on object attributes, relationships, and location annotations of bounding boxes involved in the scenes.

**VizWiz-Answering-Grounding**. The VizWiz [26] dataset is introduced to facilitate visual question answering tasks for individuals with visual impairments. In contrast to the data collection processes of previous VQA datasets, which are based on crowdsourcing and may introduce unnatural elements, VizWiz-VQA collects data directly from real users with visual impairments. This approach ensures a more accurate reflection of the genuine needs of visually impaired individuals and addresses many nuanced issues bridging theoretical design to practical implementation. VizWiz-VQA-Grounding provides ground truth grounding information for objects referred to in the question text, enabling effective training and evaluation of models in answering grounding tasks. The dataset comprises approximately 6.4k training examples, around 1.1k validation examples, and approximately 2.3k test examples.

### 4.2 Implementation Details

In the experimental implementation, the VQA module of LCV2 was utilized pre-trained models, including BLIP-VQA-large, Lens, or Git-VQA-large. Specifically, the BLIP-VQA-large model utilizes a large Vision Transformer (ViT) backbone and is trained on the VQA2.0 dataset for downstream VQA tasks. Within the Lens framework, modules responsible for describing visual content in terms of tags and attributes are implemented based on CLIP-H/14[2] and CLIP-L/14[3], respectively. The module generating multiple subtitle information utilizes the pre-trained BLIP-Caption-large model, and the reasoning module employs the Flan-T5 Large Language Model (LLM). The Git-VQA-large, as a larger version in the Git model series, undergoes fine-tuning training on the VQA v2 dataset. The LLM module in the experimental implementation, utilizing the Flan-T5-large version, possesses 780 million model parameters. The OVD/REC module of LCV2 utilizes the Grounding DINO model with REC capability, where we examine the impact of both swin-B and swin-T versions of the model on the experimental results.

Our experiments were conducted using the PyTorch framework version 2.1.2, with CUDA version 11.8, and the operating system of the computing machine was Ubuntu 18.04.6 LTS 64-bit. To assess the feasibility and effectiveness of our modular approach on lower computational and memory resources, we implemented the framework relying on a set of lower-performance hardware resources for inference and computations. The CPU model used was Intel(R) Xeon(R) Silver 4210R, and the GPU selected was the Nvidia GeForce RTX 2080Ti, with a VRAM space of only 11GB.

### 4.3 Evaluation Metrics

We assessed the VQA task performance of the models based on the accuracy of their generated responses. To evaluate the performance of our framework and baseline methods in VQA grounding for VQA tasks on the CLEVR and GQA datasets, we relied on precision (P), recall (R), and F1-score metrics to report Intersection-Over-Union (IoU) and Overlap metrics for answer grounding. However, for the VizWiz-VQA-Grounding dataset, the ground truth for answer grounding is presented in the form of image segmentation based on all points along the object boundaries. Therefore, the evaluated models, after obtaining grounding results, need to further process the predicted grounding regions and convert them into binary mask images. The IoU metric is then calculated using the pixel counts of the intersection and union of the masks.

The computation of the IoU metric can be mathematically described as follows:

$$IoU = \frac{\text{Area}(preRGN \cap gtRGN)}{\text{Area}(preRGN \cup gtRGN)} \tag{3}$$

Where Area(·) denotes the method for calculating the area of a region, $preRGN$ represents the predicted region, and $gtRGN$ stands for the ground truth region label.

The Equation (4) gives the mathematical expression of Overlap metric:

$$Overlap = \frac{\text{Area}(preRGN \cap gtRGN)}{\text{Area}(gtRGN)} \tag{4}$$

We set a specific IoU threshold to further calculate precision (P), recall (R), and F1-score, reporting the answer grounding performance of the evaluated methods.

### 4.4 Comparison with Baseline Models

We evaluated and compared the visual question-answering accuracy and grounding performance of the LCV2 framework with baseline models on the CLEVR and GQA datasets. The compared baseline models include MAC, MAC-CAP, and, on the CLEVR dataset, we additionally compared the SNMN and SNMN-Caps models. Visual images and textual information were fed into the baseline models, which generated attention maps for answer grounding. The attended regions in the attention maps were further processed using connected component analysis to extract and convert them into bounding box format, enabling a unified grounding evaluation in bbox form. The VQA module of LCV2 utilized the BLIP-VQA-large version model, and the LLM module employed the Flan-T5-large version model. Additionally, the OVD/REC module was based on the Grounding DINO implementation. We evaluated two versions of LCV2, differing primarily in the version of the Grounding DINO model used in the OVD/REC module (swin-B or swin-T version). In this section of experiments, the hyperparameters for Grounding DINO, namely box threshold and text threshold, were set to 0.25.

**GQA**. We conducted experiments on the balanced version of the GQA validation set, which provides a total of 132,062 question-answer text pairs, and ground truth bbox labels for phrase grounding of the question texts, answer texts, full answer texts, and all texts. LCV2 framework and baseline methods were evaluated based on the ground truth grounding labels for "answer" text and "all" text. In contrast to the evaluation based on "all" text grounding labels, when evaluating based on "answer" text grounding labels, the predicted answer text from the LCV2's VQA module is directly fed into the VG module to predict the grounding bbox result without being processed by the LLM module, since the nature of the given labels themselves is to assess the performance for phrase grounding related to "answer" text. Figure 5 and Figure 6 provide partial visual results of LCV2 evaluated on the GQA dataset. Table 1 presents the quantified performance comparison results between LCV2 and baseline methods, where IoU and Overlap quantitative data are obtained with a set IoU threshold of 0.5. The baseline models MAC and MAC-Caps have a time step T set to 4, as it has been reported to perform the best on this dataset. LCV2, based on the VQA module of BLIP-VQA-Large without further fine-tuning, achieves an answer accuracy of 56.6%, a 1.5% improvement over MAC-Caps. Importantly, LCV2 demonstrates a significant performance improvement in visual grounding. Based on the VG module composed of Grounding DINO swin-B version, LCV2 achieves the best performance in phrase grounding on "answer" text, while LCV2 with the VG module composed of Grounding DINO swin-T version achieves the best performance in phrase grounding on "all" text.

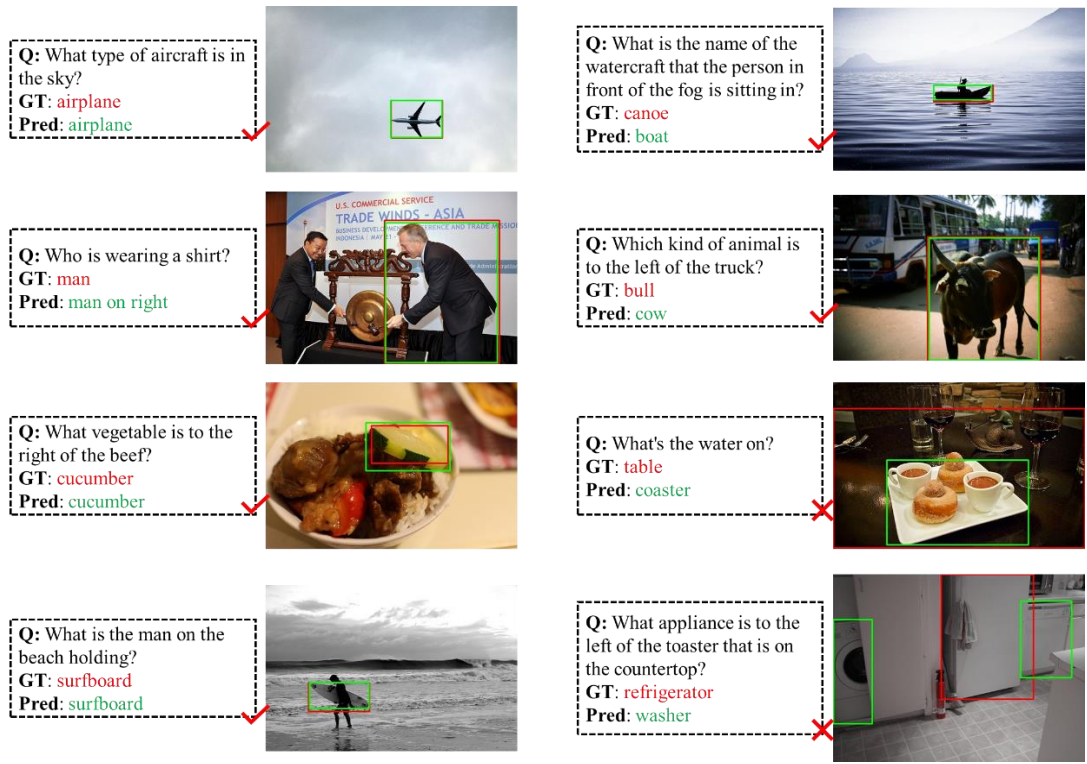

Figure 5. Visualization examples of LCV2 model's grounding results on the GQA balanced version's validation set for "answer" text, where LCV2 exclusively grounds the predicted answers in the images. Red bounding boxes represent the ground truth objects, while green boxes indicate the grounding objects predicted by LCV2. Best viewed in color.

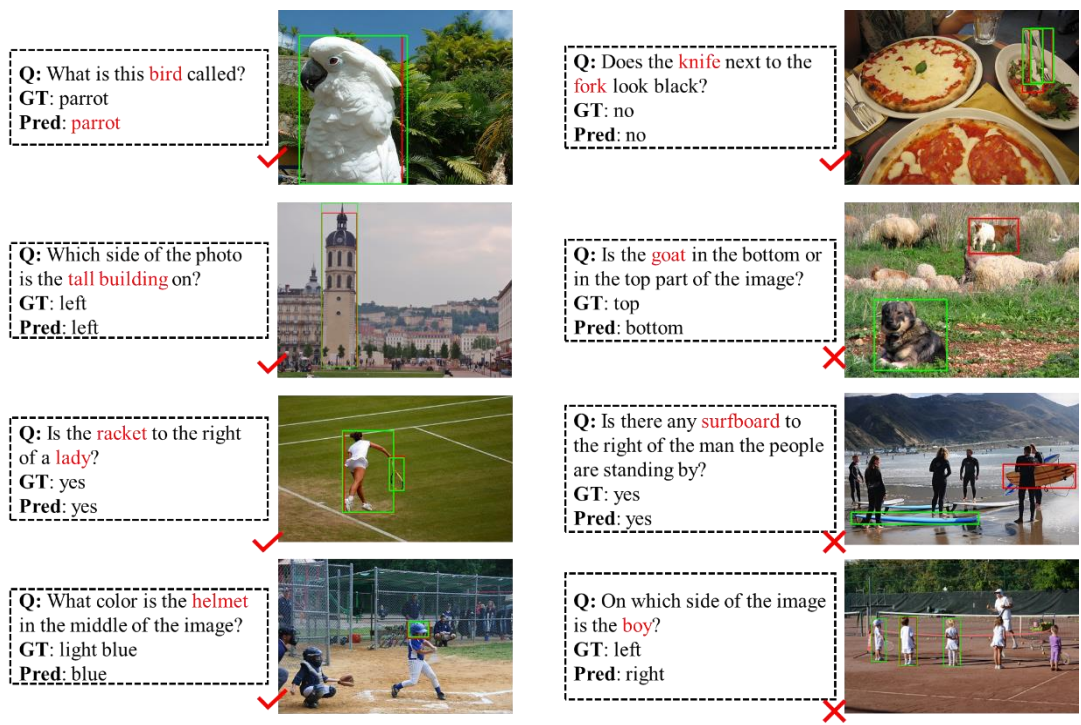

Figure 6. Visualization examples of LCV2 model's grounding results on the GQA balanced version's validation set for "all" text, where phrases marked in red indicate the grounding objects. In the images, red bounding boxes represent the ground truth labels, while green bounding boxes depict the bounding boxes predicted by LCV2. It is recommended to observe in color.

Table 1. Experimental validation of LCV2 on the GQA validation set and quantitative comparisons with baseline methods were conducted. The results of the experiments are based on the grounding of referenced objects in Answer (A) text and ALL text. In the table, LCV2 (swin-T) and LCV2 (swin-B) represent two versions of LCV2, distinguished by the OVD/REC module utilized, either constructed based on the Grounding DINO swin-T or Grounding DINO swin-B model.

| Models | Obj. | Acc. | IoU | | | Overlap | | |
|---|---|---|---|---|---|---|---|---|
| | | | Precision | Recall | F1-score | Precision | Recall | F1-score |
| MAC [12] | A | **0.571** | 0.009 | 0.045 | 0.015 | 0.056 | 0.274 | 0.093 |
| MAC-Caps [10] | | 0.551 | 0.023 | 0.119 | 0.039 | 0.120 | 0.626 | 0.201 |
| LCV2 (swin-T) | | 0.566 | 0.273 | **0.637** | 0.382 | 0.372 | 0.786 | 0.505 |
| LCV2 (swin-B) | | 0.566 | **0.323** | 0.590 | **0.417** | **0.497** | **0.805** | **0.614** |
| MAC [12] | All | **0.571** | 0.037 | 0.043 | 0.040 | 0.250 | 0.305 | 0.275 |
| MAC-Caps [10] | | 0.551 | 0.070 | 0.087 | 0.078 | 0.461 | 0.623 | 0.530 |
| LCV2 (swin-T) | | 0.566 | 0.515 | **0.707** | **0.596** | 0.751 | **0.894** | **0.816** |
| LCV2 (swin-B) | | 0.566 | **0.516** | 0.659 | 0.578 | **0.763** | 0.856 | 0.807 |

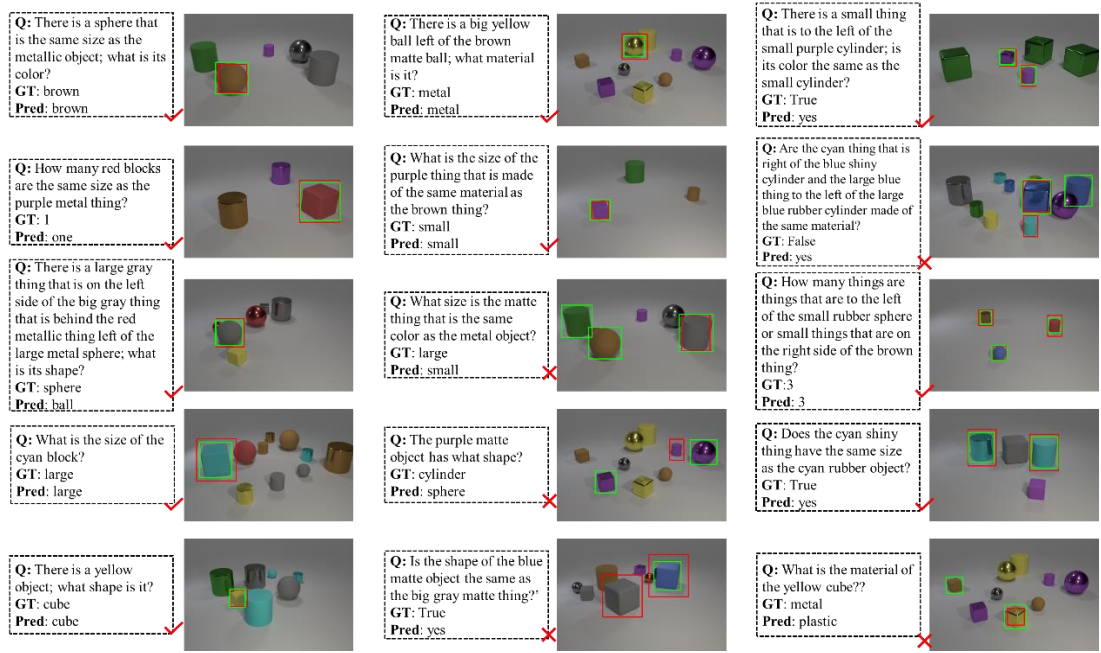

Figure 7. Visualization examples of LCV2's VQA grounding results on the CLEVR validation set are shown in several images. The images display the predicted answers generated by LCV2 alongside the ground truth answers. In the visual content, green bounding boxes represent the predictions made by LCV2, while red bounding boxes indicate the ground truth bounding boxes given by the dataset. It is recommended to view the images in color for optimal interpretation.

**CLEVR**. We also evaluated LCV2 and baseline models on the CLEVR validation set, which consists of 149,991 question-answer text pairs and provides explicit bbox labels for the specified expressions related to the questions. For baseline models MAC and MAC-Caps, we set the model's time step T to 4, 6, and 12, respectively, to compare the performance of different configurations of baseline models. For baseline models SNMN and SNMN-Caps, T was set to 9, as it has been reported to show good performance with this parameter value. We visualized the results of LCV2's inference on the CLEVR validation set, as shown in Figure 7. The quantitative results of the experiments are presented in Table 2, where the experiment data is obtained with an IoU threshold set to 0.5. LCV2, based on the

VQA module of BLIP-VQA-Large without further specific training for the CLEVR scene, achieved a relatively low answer accuracy of only 36.7%. This is attributed to the fact that the CLEVR dataset is generated based on abstract geometric shapes and does not reflect real-world scenes. The LCV2's BLIP-VQA-Large model, trained on real-world scene data, exhibits mediocre performance when dealing with such synthetic datasets. In the subsequent section of the paper, we perform fine-tuning on LCV2's VQA module on CLEVR to investigate the performance improvement. LCV2 also shows mediocre performance in visual grounding. Its IoU and overlap F1-score lag behind SNMN and SNMN-Caps, with SNMN-Caps achieving the best grounding performance in the experiment. It is noteworthy that LCV2 achieves the highest IoU and overlap recall, indicating fewer instances of false negatives. However, its IoU and overlap precision are relatively ordinary, suggesting potential instances of false positives in the detection results.

**Table** 2. Experiments were conducted on the CLEVR validation set, and the performance was quantitatively compared with baseline methods.

| Models | T | Acc. | IoU | | | Overlap | | |
| --- | --- | --- | --- | --- | --- | --- | --- | --- |
| | | | Precision | Recall | F1-score | Precision | Recall | F1-score |
| MAC [12] | 4 | 0.977 | 0.140 | 0.335 | 0.197 | 0.249 | 0.563 | 0.346 |
| MAC-Caps [10] | | 0.968 | 0.240 | 0.391 | 0.297 | 0.470 | 0.731 | 0.572 |
| MAC [12] | 6 | 0.980 | 0.126 | 0.236 | 0.164 | 0.301 | 0.524 | 0.382 |
| MAC-Caps [10] | | 0.980 | 0.290 | 0.476 | 0.361 | 0.485 | 0.798 | 0.603 |
| MAC [12] | 12 | **0.985** | 0.085 | 0.181 | 0.116 | 0.287 | 0.533 | 0.373 |
| MAC-Caps [10] | | 0.979 | 0.277 | 0.498 | 0.356 | 0.509 | 0.946 | 0.662 |
| SNMN [13] | 9 | 0.962 | 0.378 | 0.475 | 0.421 | 0.529 | 0.670 | 0.591 |
| SNMN-Caps [10] | | 0.967 | **0.506** | 0.518 | **0.512** | **0.738** | 0.781 | **0.759** |
| LCV2 (swin-t) | - | 0.367 | 0.265 | **0.577** | 0.363 | 0.418 | **0.785** | 0.545 |
| LCV2 (swin-b) | | 0.367 | 0.296 | 0.425 | 0.349 | 0.492 | 0.660 | 0.564 |

### 4.5 Impact of the VQA Module on Results

The VQA module is a critical component in LCV2, providing crucial predicted answers about visual questions. The performance of the VQA module significantly impacts the grounding performance of LCV2. Therefore, this section conducts experiments to test the impact of various pre-trained VQA models or frameworks as the VQA module in the LCV2 framework on task performance. The LLM module, consistent with Section 4.4, is still based on the frozen Flan-T5-Large model. Experiments are conducted on the same dataset range as in Section 4.4.

**GQA.** On the GQA validation set, we employed publicly available pre-trained models, namely Lens, Blip-VQA-large, and Git-VQA-large, to serve as the VQA module in LCV2. The LLM and OVD/REC modules were not replaced, resulting in three versions of the LCV2 framework. The OVD/REC module of LCV2 was constructed using the Grounding DINO swin-B version, which was pre-trained on a more extensive dataset, theoretically enabling more accurate grounding of open-world objects. Experiments were conducted based on the ground truth labels for phrase grounding using "answer" text and "all" text from the dataset. The performance changes reported by LCV2 based on different pre-trained VQA systems were evaluated using IoU and Overlap metrics. The experimental data is presented in Table 3, where the experimental data is obtained with an IoU threshold set at 0.5. LCV2's VQA module based on BLIP-VQA-Large achieved the highest answer accuracy at 56.6%.

Conversely, the LCV2 with the VQA module based on Lens demonstrated a lower accuracy of 27.8%. However, this does not necessarily imply that Lens produces incorrect answers. Instead, Lens exhibits strong answer flexibility, providing synonymous or alternative textual descriptions for ground truth answers to some extent, since Lens relies on the LLM module to extract and reason about information from the visual description module to generate predicted answers. Furthermore, Lens has certain limitations, leading to the poorest performance in accurate grounding of visual content within the constructed LCV2 in the experiment. For instance, it struggles to accurately answer questions related to the position information in the visual content. This limitation arises from the original Lens framework's visual description module, which generates tags, attributes, and captions without explicitly describing the relative or absolute positions of various objects in the visual content. Lens's predicted answers also tend to be longer, containing more phrases about visual content, even when prompted to generate short answers. This results in LCV2 generating excessive non-optimal bounding boxes for answer grounding. In contrast, LCV2 with the VQA module based on BLIP-VQA-Large achieved the best F-1 score for object grounding in visual content compared to those based on Git-VQA-Large and Lens. This superiority is attributed to the SOTA performance of BLIP-VQA-Large. We visualized the inference results, as shown in Figure 8.

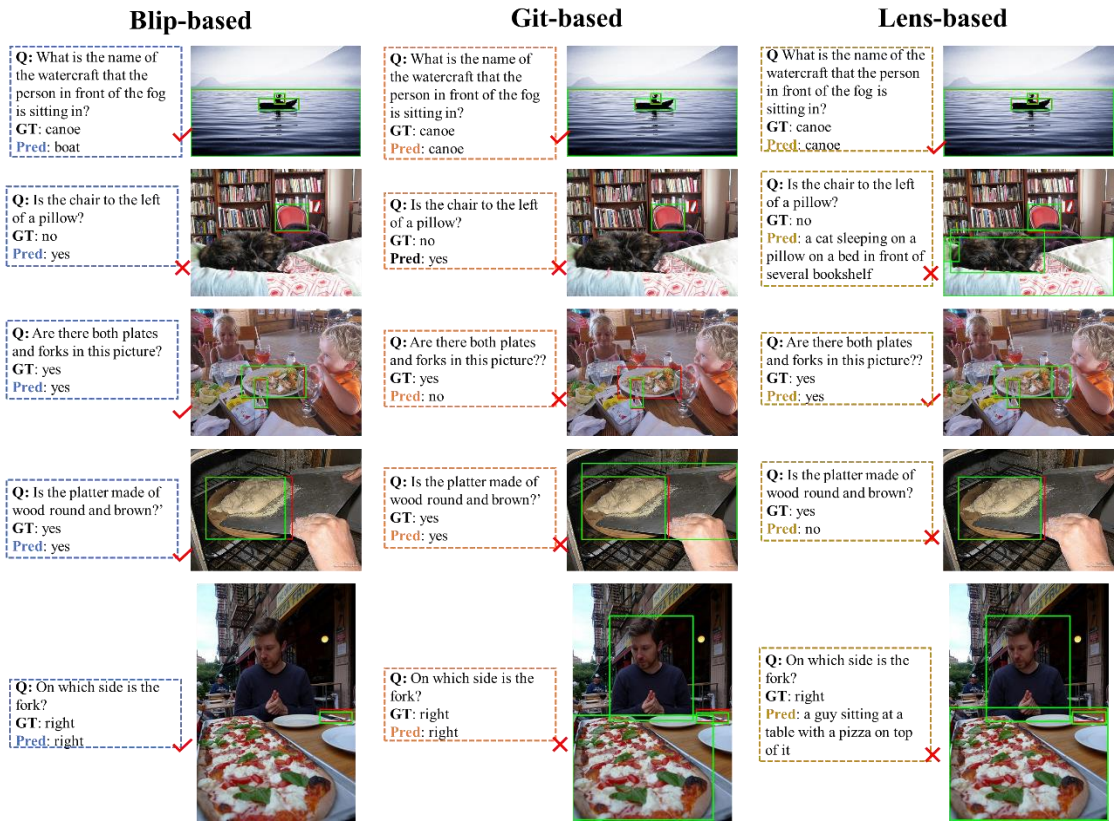

**Figure** 8. Examples of VQA grounding results on the GQA balanced validation set for LCV2 based on different VQA models. Each column represents predictions made by LCV2 based on a specific VQA model, with green and red bounding boxes indicating predicted and ground truth bounding boxes, respectively. The examples also visually demonstrate that Lens tends to provide longer prediction answers, including more objects in the visual content, leading to more non-target bounding boxes when LCV2 performs open-world detection.

**CLEVR**. Since the visual scenes in CLEVR are not captured from the real world but are composed of abstract three-dimensional geometric shapes generated by computers, the publicly available pre-trained BLIP-VQA-Large model, pre-trained on VQA2.0 (real-world visual content), exhibits relatively modest performance on the CLEVR VQA task. To enhance the model's performance, we conducted efficient fine-tuning of the BLIP-VQA-Large model on the CLEVR training set using the LoRA method, whose mathematical description is introduced in Section 3.1. The CLEVR training set provides 70k images and 699,989 pairs of question-answer text. For efficient fine-tuning, trainable bypasses were added to the Q-matrix and V-matrix of the transformer framework in BLIP-VQA-Large. The rank of the trained matrices was set to 16, the scaling factor to 16, dropout ratio to 0.1, and no bias parameters were trained. The number of parameters retrained in the BLIP-VQA-Large model amounted to 2,359,296, constituting 0.610% of the total model parameters. Considering the available hardware resources, the training time for one epoch was approximately 8 hours, and we conducted a simple 5-epoch fine-tuning to obtain the BLIP-VQA-Large-finetuned model on CLEVR. The experimental results are presented in Table 4. Utilizing the BLIP-VQA-Large model fine-tuned on the CLEVR dataset as the VQA module in the LCV2 framework led to a significant increase in answer accuracy, reaching 0.773. Correspondingly, due to the improvement in answer accuracy, the accuracy of visual content grounding also increased. LCV2 constructed with the OVD/REC module based on Grounding DINO swin-T achieved the best IoU F1-score, quantified at 0.374, while LCV2 based on Grounding DINO swin-B achieved the best Overlap F1-score, reaching 0.577. The experimental results further demonstrate that modules within the LCV2 framework can be effectively fine-tuned to enhance adaptability to specific datasets and domains.

Table 3. Validating the impact of LCV2 based on different VQA models on performance outcomes on the GQA validation set. Three versions of LCV2 are obtained based on the different VQA modules constructed. The experimental results are based on the grounding of referenced objects in Answer (A) text and all text.

| Models | Obj. | Acc. | IoU | | | Overlap | | |
|---|---|---|---|---|---|---|---|---|
| | | | Precision | Recall | F1-score | Precision | Recall | F1-score |
| LCV2 (Blip-L) | | **0.566** | **0.323** | **0.590** | **0.417** | **0.497** | **0.805** | **0.614** |
| LCV2 (lens) | A | 0.278 | 0.261 | 0.505 | 0.345 | 0.491 | 0.795 | 0.607 |
| LCV2 (Git) | | 0.518 | 0.292 | 0.545 | 0.380 | 0.463 | 0.770 | 0.579 |
| LCV2 (Blip-L) | | **0.566** | **0.516** | 0.659 | **0.578** | **0.763** | 0.856 | **0.807** |
| LCV2 (lens) | All | 0.278 | 0.414 | 0.612 | 0.494 | 0.692 | **0.858** | 0.776 |
| LCV2 (Git) | | 0.518 | 0.506 | 0.649 | 0.568 | 0.756 | 0.852 | 0.801 |

Tabel 4. The experiment mainly validates the impact of two versions of the BLIP-VQA-Large model, one fine-tuned using LoRA on the CLVER training set and the other without fine-tuning, as the VQA module for LCV2 on the CLEVR validation set.

| Models | Acc. | IoU | | | Overlap | | |
|---|---|---|---|---|---|---|---|
| | | Precision | Recall | F1-score | Precision | Recall | F1-score |
| LCV2 (swin-t) | 0.367 | 0.265 | 0.577 | 0.363 | 0.418 | 0.785 | 0.545 |
| LCV2 (swin-b) | 0.367 | 0.296 | 0.425 | 0.349 | 0.492 | 0.660 | 0.564 |
| LCV2 (finetuned-Blip, swin-t) | **0.773** | 0.273 | **0.596** | **0.374** | 0.424 | **0.801** | 0.554 |
| LCV2 (finetuned-Blip, swin-b) | **0.773** | 0.312 | 0.425 | 0.360 | 0.512 | 0.662 | **0.577** |

### 4.6 Experimental Setup and Analysis on the VizWiz Answer Grounding

An important and practically significant application scenario of the VQA grounding task is to assist individuals with visual impairments in answering to simple questions about visual content and providing explicit grounding information for referenced objects within the visual context. The VizWiz Answer Grounding dataset is derived from real-world scenarios and requirements of visually impaired individuals, serving as a benchmark for evaluating the applicability performance of relevant methods in the context of the answer grounding task for individuals with visual impairments.

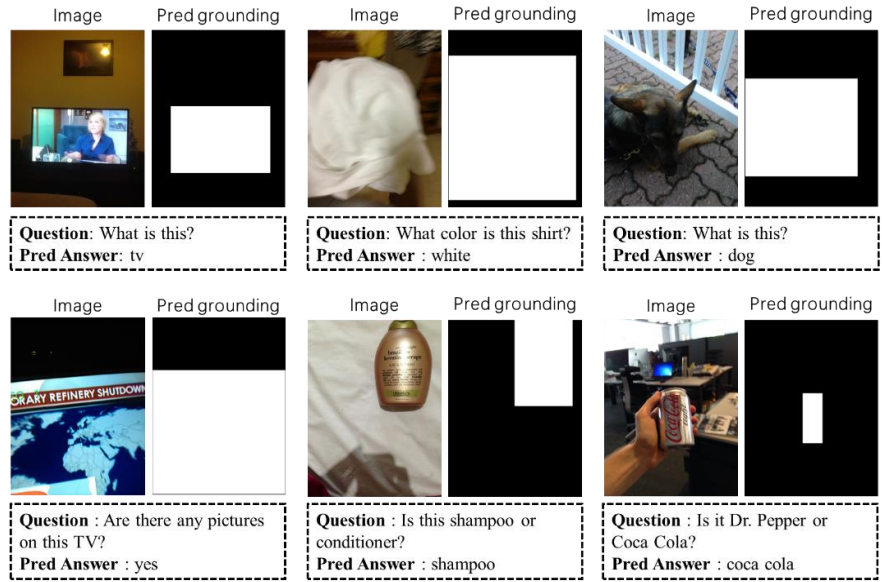

**Figure** 9. Visualization of several prediction results of the LCV2 model on the VizWiz Answer Grounding test set. The LCV2's predicted answer grounding is presented in the form of binary mask images.

In this section, we deploy LCV2 on the test set of the VizWiz Answer Grounding dataset, which comprises 2373 test examples. We actively participated in and submitted entries for the online challenge hosted by EvalAI: test-standard2023-vizwiz-VQA-Grounding. The evaluation results are reported using the average IoU metric. We compare LCV2 with some SOTA methods, and the experimental data is presented in Table 5. The evaluated LCV2 framework is composed of BLIP-VQA-Large fine-tuned on the VizWiz Answer Grounding training set, the large language model Flan-T5, and the Grounding DINO swin-B. The results of the VQA and Answer Grounding tasks inferred by LCV2 are visualized in Figure 9. The answer grounding results are transformed into binary mask images, where the foreground or the internal region of the grounding is represented by white, and the background or the external region is represented by black.

Table 5. Evaluate the performance of LCV2 on the VizWiz Answer Grounding test set and compare it with some state-of-the-art (SOTA) methods, with results reported in terms of the IoU metric.

| Year | Team | IoU |
| --- | --- | --- |
| 2022 | Aurora (ByteDance & Tianjin University) | 0.71 |
|  | hsslab_inspur | 0.70 |
|  | Pinkiepie | 0.33 |
|  | binggo | 0.08 |
| 2023 | UD VIMS Lab (EAB) | 0.74 |
|  | MGTV_Baseline | 0.72 |
|  | DeepBlue_AI | 0.69 |
|  | USTC | 0.46 |
|  | ours | 0.43 |

However, our method exhibits inferior performance on this dataset, with a significant contributing factor being that our framework provides grounding in the form of bbox. The ground truth labels provided by VizWiz Answer Grounding, on the other hand, are in a more fine-grained image segmentation format, composed of points along all edges of the answer grounding. This discrepancy results in our method, even when providing accurate answer grounding information, having a negative impact on the experimental result of answer grounding due to the non-fine-grained nature of the grounding results. In future work, replacing the OVD/REC module in our framework with a model capable of providing grounding information in segmentation form could further enhance the method's performance on answer grounding tasks.

Table 6. Validate the performance variation of LCV2 on the VizWiz Answer Grounding dataset based on different VQA models.

| Models | IoU |
| --- | --- |
| LCV2 (lens-swinB) | 0.424 |
| LCV2 (Gittext-swinB) | 0.405 |
| LCV2 (BlipL-swinB) | 0.425 |
| LCV2 (BlipL-swinT) | 0.417 |
| LCV2 (FineTunedBlipL-swinB) | **0.430** |

To investigate the impact of different publicly available pretrained VQA models constituting the VQA module of LCV2 on task performance, comparative experiments were conducted. We employed Lens, GittextVQA, BLIP-VQA-Large model, and FineTuned BLIP-VQA-Large model as the VQA module of LCV2, observing the influence on the final answer grounding's IoU quantification results. It is worth mentioning that the utilized GittextVQA is based on the Git-VQA-large model fine-tuned on the TextVQA dataset [50], possessing the ability to recognize text and symbols in visual content. The experimental data are presented in Table 6. LCV2 constructed with BLIP-VQA-Large and Grounding DINO swin-B achieved an IoU of 0.425, showing a certain performance improvement over LCV2 constructed with BLIP-VQA-Large and Grounding DINO swin-T in the Answer Grounding task, where the latter obtained an IoU metric of 0.417, mainly attributed to the superior performance of Grounding DINO swin-B in open-world object localization tasks. LCV2 with the VQA module based on the GittextVQA model exhibited the poorest performance in the experiment, with an IoU of 0.405, possibly due to the inferior visual question-answering performance of Git-VQA-Large compared to BLIP-VQA-Large and the original Lens. LCV2's VQA module, based on the BLIP-VQA-Large model fine-tuned on the VizWiz Answer Grounding training set, achieved the best performance. The experimental results further indicate that fine-tuning contributes to enhancing the adaptability of the VQA module to the VizWiz Answer Grounding dataset. However, as these versions of LCV2 provide visual answer grounding results in the form of bounding boxes, their performance is not particularly outstanding. Future work should focus on the ability of the OVD/REC module to provide more fine-grained image segmentation forms of grounding to effectively improve the model's performance in answer grounding tasks.

## 5. Conclusion

We propose the LCV2 modular approach to accomplish the VQA grounding task, aiming to answer textual questions about visual content while providing explicit visual cues. This approach holds the potential for applications in visual navigation for the visually impaired and broader human-computer interaction scenarios. Our method introduces a pretraining-free general framework that allows deployment and implementation on lower computational power and storage resources, enabling the realization of VQA grounding tasks with reduced hardware demands. The method leverages a frozen pretrained Large Language Model (LLM) to facilitate the connection and transformation of information between the VQA and OVD/REC systems, constructing an integrated and flexible unified framework. The modules within the framework can be easily adopted and replaced with state-of-the-art pretrained models and methods in the community without considering their pretraining processes and heterogeneous networks. Furthermore, the modules in LCV2 can be fine-tuned to enhance the method's applicability and specialization to specific datasets.

We validate our approach on benchmark datasets, including GQA and CLEVR, and compare its performance with several baseline methods. The experiments report the IoU and Overlap metrics for visual grounding based on precision (P), recall (R), and F1-score. Extensive experiments demonstrate that our method achieves competitive performance on these datasets. Additionally, we test our model on the VizWiz Answer Grounding dataset to evaluate its performance in real-world applications, specifically in visual answer grounding for individuals with visual impairments, validating the effectiveness of our proposed approach.

The next step in the work, the model framework's modules can be replaced with publicly available pre-trained models with larger parameter sizes and superior performance, given the availability of more powerful computational and storage resources. Additionally, considering a finer-grained Answer Grounding system to serve as the OVD/REC module could enhance LCV2's precision in visual grounding, for instance, with the ability to provide localization information in the form of image segmentation. Furthermore, LCV2 could be extended to a broader range of interesting tasks in other domains, such as question-answering and grounding in audio or video, yielding valuable insights. This expansion would further facilitate the integration and coordination of LLM and multimodal models, exploring intriguing directions for LLM applications in the community.